\pdfoutput=1

\documentclass[11pt]{article}

\usepackage{acl}

\usepackage{times}
\usepackage{latexsym}

\usepackage[T1]{fontenc}

\usepackage[utf8]{inputenc}

\usepackage{microtype}

\usepackage{graphicx}
\usepackage{enumerate}
\usepackage{booktabs}

\newcommand*{\affaddr}[1]{#1} 
\newcommand*{\affmark}[1][*]{\textsuperscript{#1}}
\newcommand*{\email}[1]{\texttt{#1}}

\title{How Relevant is Selective Memory Population in \\ Lifelong Language Learning?}

\author{%
Vladimir Araujo\affmark[1,2], Helena Balabin\affmark[1], Julio Hurtado\affmark[3], Alvaro Soto\affmark[2], Marie-Francine Moens\affmark[1]\\
\affaddr{\affmark[1]KU Leuven},
\affaddr{\affmark[2]Pontificia Universidad Católica de Chile},
\affaddr{\affmark[3]University of Pisa}\\
\email{vgaraujo@uc.cl},
\email{helena.balabin@kuleuven.be},
\\
\email{julio.hurtado@di.unipi.it},
\email{asoto@ing.puc.cl},
\email{sien.moens@kuleuven.be}
}

\begin{document}
\maketitle
\begin{abstract}


Lifelong language learning seeks to have models continuously learn multiple tasks in a sequential order without suffering from catastrophic forgetting. State-of-the-art approaches rely on sparse experience replay as the primary approach to prevent forgetting. Experience replay usually adopts sampling methods for the memory population; however, the effect of the chosen sampling strategy on model performance has not yet been studied. In this paper, we investigate how relevant the selective memory population is in the lifelong learning process of text classification and question-answering tasks. We found that methods that randomly store a uniform number of samples from the entire data stream lead to high performances, especially for low memory size, which is consistent with computer vision studies.

\end{abstract}

\section{Introduction}

While humans learn throughout their lifetime, current deep learning models are restricted to a bounded environment, where the input distribution is fixed.
When those models are sequentially learning new tasks, they suffer from catastrophic forgetting \citep{mccloskey1989catastrophic, ratcliff1990connectionist} because the input distribution changes.

Several methods have been proposed to address catastrophic forgetting, mainly for computer vision (CV) \citep{9349197} and few others for natural language processing (NLP) \citep{biesialska-etal-2020-continual}. In both, one of the prominent approaches is experience replay with episodic memory \citep{10.1162/neco_a_01433}, which aims to store previously seen training examples and later use them to perform gradient updates while training on new tasks.

In the experience replay approach, random sampling is the de facto method for the memory population, as it has shown good results in CV \citep{chaudhry_tiny_2019,8954008,10.1007/978-3-030-58598-3_28}. In contrast, other works have shown that memory selection is relevant for deep reinforcement learning \citep{10.5555/3504035.3504439}, image classification \citep{10.1007/978-3-030-01252-6_33,sun2022informationtheoretic}, and analogical reasoning \citep{hayes_selective_2021}. However, no previous work has explored NLP tasks, which raises the question of whether memory selection is necessary for lifelong language learning.

In this paper, we adopt and evaluate seven memory population methods under a lifelong language learning setup with sparse experience replay. We conducted experiments with text classification and question answering tasks. 
We find that methods that obtain memory with a random sample from the global data distribution for text classification provide the best results in both high and low memory regimes. Conversely, for the question answering task, a method that provides a balanced memory composition per task performs better.





\section{Related Work}

\paragraph{Lifelong Learning in NLP.}
Rather than training a language model on a fixed dataset, lifelong (continual) language learning setups consist of a stream of tasks (e.g., text classification). In this setup, a model aims to retain the most relevant information to prevent catastrophic forgetting.
Existing approaches for NLP include purely replay-based methods \citep{NEURIPS2019_f8d2e80c,han-etal-2020-continual,Araujo_2022_CVPR}, meta-learning based methods \citep{wang-etal-2020-efficient,holla2020lifelong} and generative replay-based methods \citep{sun_lamol_2020,sun-etal-2020-distill}.


\paragraph{Memory Selection in Lifelong Learning.}
Several strategies have been proposed to store and select the most relevant training examples in memory. 
Early work has shown that reservoir sampling prevents catastrophic forgetting in lifelong reinforcement learning \citep{10.5555/3504035.3504439} and supervised learning \citep{chaudhry_tiny_2019} with limited memory.
More recent works have explored criteria-based selection methods, showing that maximum-loss examples are helpful for analogical reasoning \citep{hayes_selective_2021} and gradient-based \citep{10.5555/3454287.3455345} or information-theoretic \citep{sun2022informationtheoretic} selection for image classification.



\section{Lifelong Language Learning Setup}
\label{sec:lll}

We consider the lifelong language learning setting proposed by \citet{NEURIPS2019_f8d2e80c}, in which a model learns multiple tasks in sequential order from a stream of training examples\footnote{We use an available implementation of this setup: \href{https://github.com/vgaraujov/LLL-NLP}{https://github.com/vgaraujov/LLL-NLP}}. 
In this setup, each example is only allowed to be viewed once.


This setup adopts sparse experience replay, which performs a gradient update at a certain interval during training. We leverage this method, as \citet{NEURIPS2019_f8d2e80c} have shown that a sparse 1\% rate of replaying to learning new examples is sufficient for lifelong language learning. 


This setting also includes local adaptation \citep{sprechmann2018memorybased}, which is a process that retrieves K-nearest neighbors examples from memory to update model parameters used to predict a particular test example.
However, recent works have tried to reduce its use \citep{wang-etal-2020-efficient} or even avoid it \citep{holla2020lifelong} because it significantly slows down the inference speed. 
We do not use this mechanism in our main experimentation because our goal is to analyze the effect of selective memory on the generalization of the model. Nevertheless, Section~\ref{sec:results} briefly shows how resulting memory composition influences local adaptation.

\section{Selective Episodic Memory}
For the previously described lifelong learning setup, we extend a replay model (see Section \ref{model}) with the following seven memory population methods:

\paragraph{Naive Random.} A basic method for memory population. It samples a percentage of elements of each task. In our experiments, the percentage value is the same as the memory capacity, and we sample the elements on the fly from the current batch. 

\paragraph{Reservoir.} A reservoir \citep{10.1145/3147.3165} allows sampling elements from a stream without knowing how many elements to expect. 
It samples each element with a probability  $\frac{M}{N}$ where $N$ is the number of elements observed so far and $M$ is the memory size. This way, it acts randomly to maintain a uniform sample from the already seen stream.

\paragraph{Ring Buffer.} Similar to \citet{NIPS2017_f8752278}, this method allocates $\frac{M}{C}$ elements for each class $C$ of the task in memory. The strategy is a FIFO buffer, so the memory is always filled with the latest task observations. If the total number of classes is unknown, the value of $M$ is gradually reduced as new tasks are observed.

\paragraph{Surprise.} Unexpected events have been shown to influence episodic memory in humans \citep{Cheng2008}. 
One way to measure surprise is by computing the entropy of the output distribution of an input batch.
Analogous to \citet{10.5555/3504035.3504439}, we use the time difference between the current entropy value and that of the previous batch to sample high-surprise elements.


\paragraph{Minimum Margin.} Similar to \citet{hayes_selective_2021}, who introduced a margin-based method for CV replay models, we define the margin as the difference between the probability of the true class and the probability of the other most likely class. We store the most uncertain examples, that is, those with the smallest margin for which the probability of the true class is only marginally different from the probability of the other most likely class.

\paragraph{Maximum Loss.} Analogous to the previous strategy, the maximum loss strategy aims to store samples with high uncertainty. However, this time it is based on storing samples with a high loss value \citep{hayes_selective_2021}. Here, we slightly modify the strategy by evaluating the loss for an entire batch, therefore storing and overriding whole batches in memory.


\paragraph{Mean of Features (MoF).} Similar to \citet{rebuffi_icarl_2017, chaudhry_tiny_2019}, we calculate the average feature vector based on averaging the final \texttt{[CLS]} representations in memory for a given class. If the representation of an input example has a smaller distance to its average feature vector than the entry in the memory with the largest distance to the average, we store the new incoming example and update the respective average feature vector.

\begin{table*}[]
\small
\centering
\vspace{-0.5cm}
\begin{tabular}{lccccccc}
\toprule
Order & N. Random    & Reservoir    & Ring Buffer   & Surprise     & Min. Margin  & Max. Loss & MoF  \\
\midrule
\multicolumn{8}{c}{Text Classification (Accuracy)}                                                         \\
\midrule
i.    & 70.88$\pm$1.22 & 69.54$\pm$5.99 & 68.36$\pm$3.61 & 53.74$\pm$1.83 & 71.40$\pm$0.83 & 56.59$\pm$1.61 & 60.34$\pm$7.39  \\
ii.   & 72.17$\pm$0.41 & 73.41$\pm$1.14 & 74.32$\pm$0.35 & 69.40$\pm$2.14 & 71.68$\pm$1.32 & 70.82$\pm$2.62 & 65.62$\pm$4.87 \\
iii.  & 65.37$\pm$1.32 & 67.79$\pm$1.34 & 65.13$\pm$2.29 & 63.00$\pm$2.44 & 63.35$\pm$0.69 & 67.64$\pm$0.96 & 56.98$\pm$2.46 \\
iv.   & 72.72$\pm$0.79 & 73.32$\pm$0.89 & 69.99$\pm$2.35 & 57.46$\pm$2.97 & 72.29$\pm$1.02 & 59.63$\pm$2.25 & 63.30$\pm$1.31 \\
\midrule
avg.  & 70.29$\pm$0.94 & 70.99$\pm$2.34 & 69.45$\pm$2.15 & 60.90$\pm$2.35 & 69.68$\pm$0.96 & 63.67$\pm$1.86 & 61.56$\pm$4.01 \\
\midrule \midrule
\multicolumn{8}{c}{Question Answering (F1 score)}                                                          \\
\midrule
i.                                      & 59.32$\pm$1.12 & 59.34$\pm$0.73 & 59.12$\pm$0.63 & 61.24$\pm$0.08 & 59.24$\pm$1.03 & 59.40$\pm$1.06 & 59.42$\pm$0.42 \\
ii.                                     & 58.40$\pm$1.22 & 58.99$\pm$0.53 & 59.38$\pm$0.26 & 59.51$\pm$0.44 & 58.48$\pm$0.67 & 59.62$\pm$0.64 & 57.06$\pm$0.95 \\
iii.                                    & 52.95$\pm$1.44 & 53.47$\pm$0.51 & 54.61$\pm$0.78 & 50.10$\pm$0.64 & 53.02$\pm$0.64 & 44.77$\pm$1.04 & 50.37$\pm$3.81 \\
iv.                                     & 60.56$\pm$0.76 & 60.03$\pm$0.18 & 60.49$\pm$0.62 & 61.00$\pm$0.39 & 59.93$\pm$0.69 & 60.16$\pm$0.48 & 59.69$\pm$0.47 \\
\midrule
avg.                                    & 57.81$\pm$1.13 & 57.96$\pm$0.48 & 58.40$\pm$0.57 & 57.96$\pm$0.39 & 57.67$\pm$0.76 & 55.99$\pm$0.80 & 56.63$\pm$1.41 \\
\bottomrule
\end{tabular}
\vspace{-0.15cm}
\caption{
Summary of results for text classification and question answering using sparse experience replay and selective episodic memory population approaches.
We report the mean accuracy or F1 score as well as the respective standard deviation across five runs with different random seeds.
}
\label{tab:results_TC_QA}
\vspace{-0.3cm}
\end{table*}


\section{Experimental Setup} \label{experiments}
\paragraph{Datasets.}
We adopt the evaluation methodology and datasets proposed by \cite{NEURIPS2019_f8d2e80c}.

For text classification, we use five datasets from \citep{NIPS2015_250cf8b5}: AGNews classification, Yelp sentiment analysis, Amazon sentiment analysis, DBPedia article classification and Yahoo questions and answers categorization. Both sentiment analysis tasks share the same labels. In total, we obtain 575,000 training and 38,000 test examples with 33 classes from all datasets using four task orders:

\begin{enumerate}[(i)]
    \vspace{-0.20cm}
    \small
    \itemsep-0.25em 
    \item Yelp → AGNews → DBPedia → Amazon → Yahoo
    \item DBPedia → Yahoo → AGNews → Amazon → Yelp
    \item Yelp → Yahoo → Amazon → DBpedia → AGNews
    \item AGNews → Yelp → Amazon → Yahoo → DBpedia
    \vspace{-0.20cm}
\end{enumerate}

For question answering, we use the following three datasets: SQuAD 1.1 \citep{rajpurkar-etal-2016-squad}, QuAC \citep{choi-etal-2018-quac}, and TriviaQA \citep{joshi-etal-2017-triviaqa}. The latter has two sections, Web and Wikipedia, which we consider separate datasets. We obtain 60,000-90,000 training and 7,000-10,000 validation examples per task, and use the following task orders:


\begin{enumerate}[(i)]
    \vspace{-0.20cm}
    \small
    \itemsep-0.25em 
    \item QuAC→ TrWeb → TrWik → SQuAD
    \item SQuAD → TrWik → QuAC→ TrWeb
    \item TrWeb → TrWik → SQuAD → QuAC
    \item TrWik → QuAC→ TrWeb → SQuAD
    \vspace{-0.20cm}
\end{enumerate}

\paragraph{Model and Memory Details.} \label{model}
We use a pre-trained BERT model augmented with an episodic memory to perform sparse experience replay.
For text classification, we use the \texttt{[CLS]} token and a classifier to predict the class. 
For question answering, we apply two linear transformations to the BERT outputs for each token to predict the probability that the token is the start/end position of an answer.
We implement the model using the huggingface library \citep{wolf-etal-2020-transformers}.
To train the model for both text classification and question answering, we use the Adam optimizer with a learning rate of $3e^{-5}$ and a training batch of size 32.
We use the BERT base version and its default vocabulary in our experiments.

\begin{table}
\small
\centering
\begin{tabular}{cc}
\toprule
\textbf{Approach}    & \textbf{Runtime} \\ \midrule 
N. Random      & 45m    \\ 
Reservoir   & 49m    \\ 
Ring Buffer & 51m    \\ 
Surprise    & 1h 27m    \\ 
Min. Margin & 1h 20m    \\ 
Max. Loss   & 46m    \\ 
MoF         & 2h 16m   \\
\bottomrule
\end{tabular}
\caption{Training time comparison of all seven memory population approaches for text classification, based on running task order (i) with one random seed on an NVIDIA GeForce RTX 3090.}
\label{tab:runtime}
\vspace{-0.4cm}
\end{table}

The episodic memory is a buffer that stores veridical inputs and labels using the memory population methods mentioned above. 
We use an experience replay rate of 1\% and memory capacity of 10\%, which \citet{NEURIPS2019_f8d2e80c} showed to be enough for good results (see Section \ref{sec:results} for additional experiments with varying memory sizes). 
We determine the memory capacity percentage based on the total size of the datasets.
The retrieval process is performed randomly from the memory with a uniform probability.
Regarding population for question answering task, all methods based on the number of classes were adapted to work based on the number of tasks. This is because question answering is a span prediction task with no classes.


\begin{figure}
  \centering
  \includegraphics[width=0.45\textwidth]{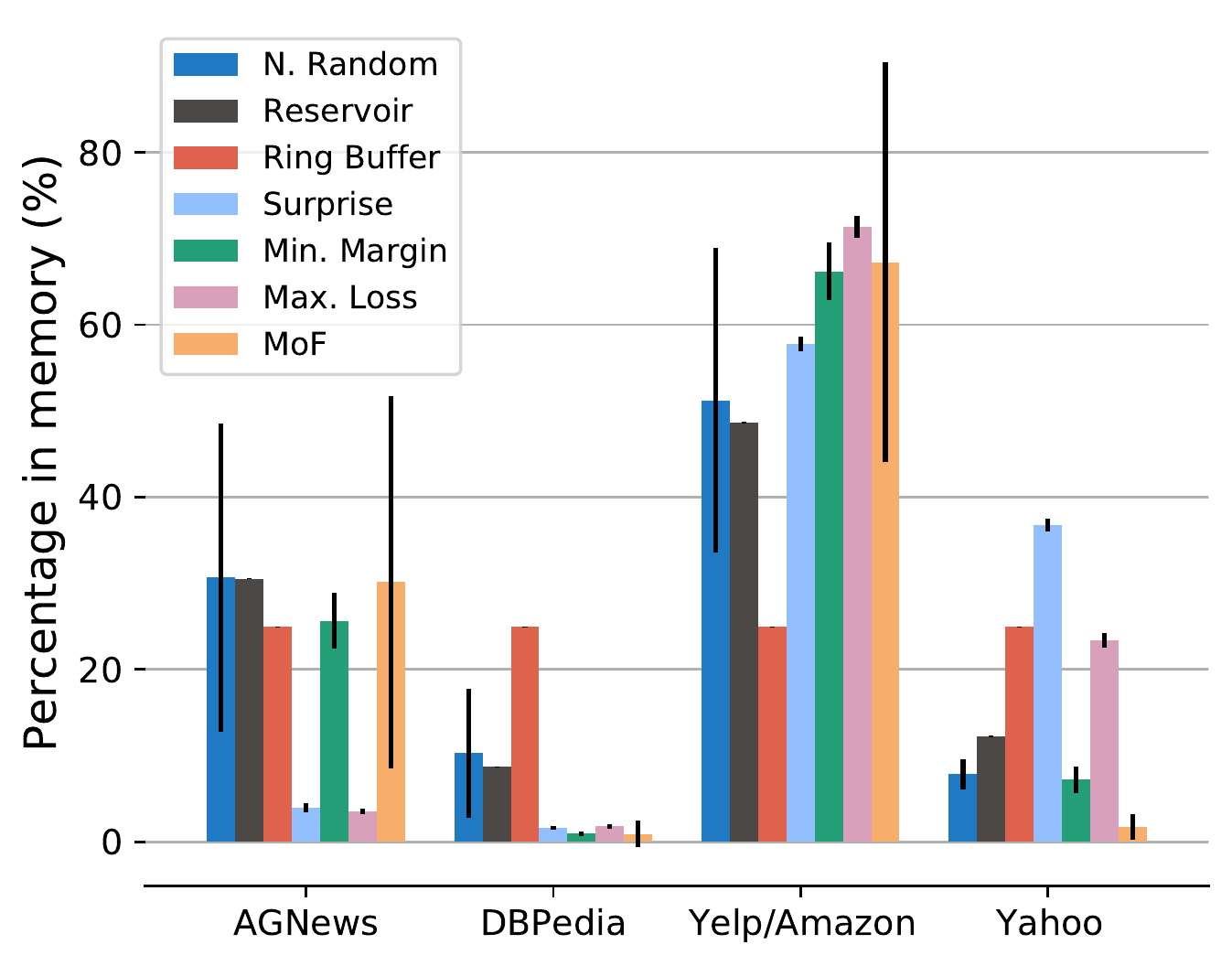}
  \vspace{-0.25cm}
  \caption{Percentage of samples in memory per task after training the model for text classification. Each color represents a different population method.}
  \label{fig:composition}
  \vspace{-0.4cm}   
\end{figure}

\section{Results}
\label{sec:results}
\paragraph{Performance.} Text classification and question answering results are shown in Table~\ref{tab:results_TC_QA}, in the upper and lower sections respectively. 
For text classification, on average, \textit{Reservoir} proved to be the best performing approach, with the \textit{Naive Random} memory placing second. Overall, the standard deviations 
tend to have larger values than the differences across approaches in many cases. 




For the question answering problem, \textit{Ring Buffer} memory performed best. Next, the \textit{Naive Random}, \textit{Reservoir}, \textit{Surprise} and \textit{Min. Margin} methods performed similarly. Compared to the text classification results, the differences in average performance across models and the standard deviations are  substantially smaller. 
This difference could be due to the more homogeneous nature of the question answering tasks (i.e., start and end span predictions), contrary to the heterogeneous set of classes used in a stream of text classification tasks.



Overall, the \textit{Max. Loss} and \textit{Surprise} method results in lower returns, which is inconsistent with previous findings from CV \citep{hayes_selective_2021, 10.5555/3504035.3504439}. For the \textit{MoF} approach, we were not able to replicate the improvement in performance \citep{chaudhry_tiny_2019} in this NLP-specific application. We suspect that this is caused by the unsuitability of the \texttt{[CLS]} token for semantic similarity purposes \citep{reimers_sentence-bert_2019}.
Finally, \textit{Reservoir} leads to the best results as it maintains a random sample over a global distribution that is not known in advance. This supports previous work on CV \citep{chaudhry_tiny_2019}, which defaults to the reservoir sampling due to its simplicity and efficiency.

We were able to confirm that the \textit{Reservoir} and \textit{Naive Random} methods are indeed the most efficient in terms of their required training time, together with \textit{Max. Loss} and \textit{Ring Buffer} (see Table \ref{tab:runtime}). Notably, \textit{MoF} is the most inefficient of the presented approaches, likely due to frequent updates of the average feature vector.

\begin{figure}
  \centering
  \includegraphics[width=0.47\textwidth]{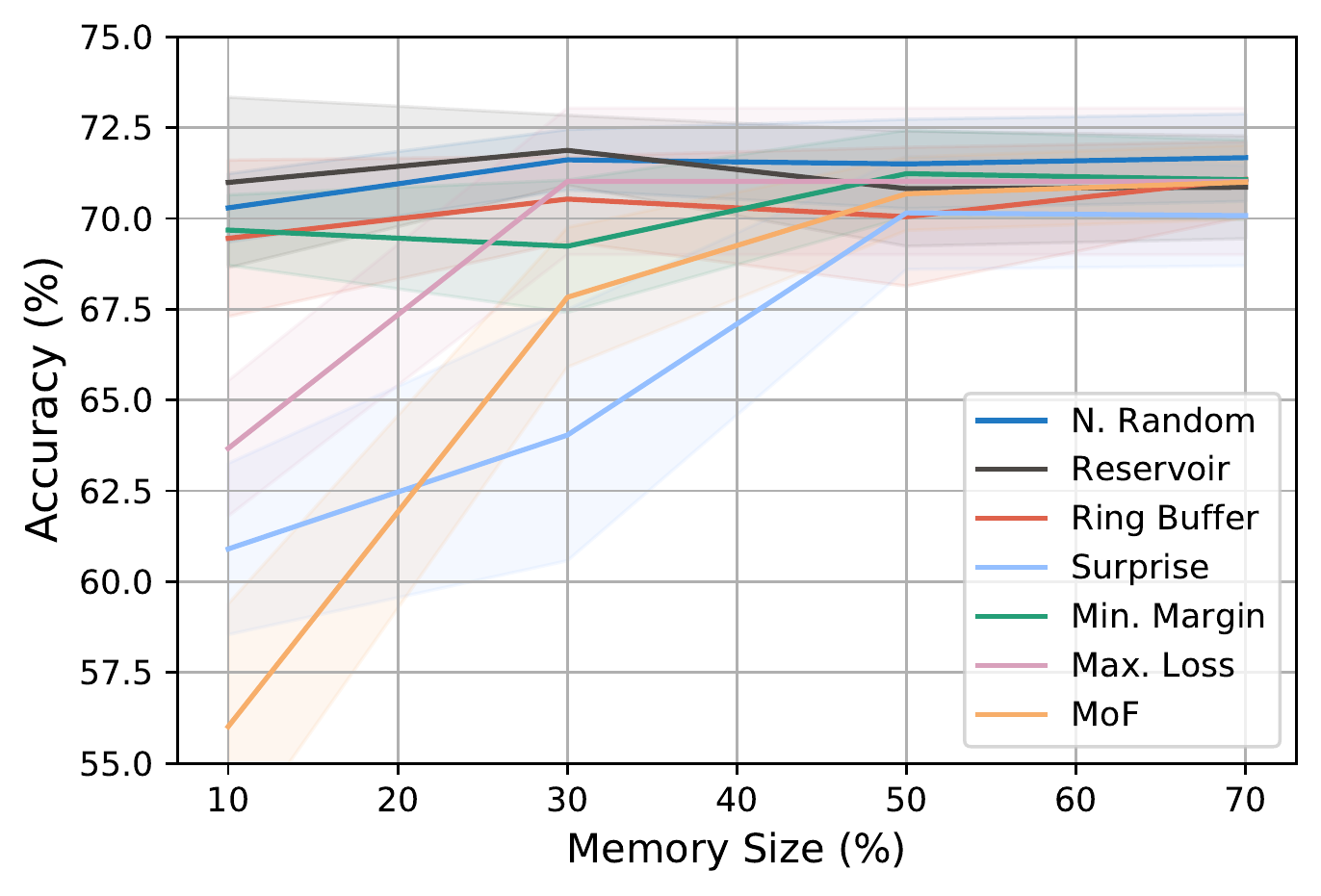}
  \vspace{-0.25cm}
  \caption{Sparse replay model performance for each population method with 10\% to 70\% memory size.}
  \label{fig:memsz}
  \vspace{-0.35cm}   
\end{figure}

\paragraph{Resulting Memory Composition.}


Figure~\ref{fig:composition} depicts the resulting memory composition after training the model for text classification tasks.
Specifically, it shows the percentage of items in memory per task normalized by the number of classes for all population methods.
We join the Yelp and Amazon datasets because of their shared classes, resulting in an overpopulation in memory.
As expected, \textit{Ring Buffer} results in a balanced number of samples. Regarding the best performing methods, \textit{Naive Random} and \textit{Reservoir}, we observe similar behaviors, possibly explaining their similar performance. However, \textit{Reservoir} better balances the number of instances per task, limiting the high number of examples stored for Yelp/Amazon.


Furthermore, certain methods result in an extremely imbalanced memory composition, which tends to hurt performance \citep{pmlr-v119-chrysakis20a}.
For instance, \textit{Surprise} and \textit{Max. Loss} are biased towards the last seen tasks (as they produce high surprise or loss), reducing the population of initial ones. Also, \textit{MoF} stores nearby items, limiting the storage of previously unseen task instances.


\paragraph{Memory Size Impact.} Figure~\ref{fig:memsz} shows the performance for text classification for memory sizes between 10\% and 70\%. Most methods do not result in a performance advantage when the memory size increases, and between 50\% and 70\% capacity, all approaches tend to perform similarly. 

However, methods with an extremely imbalanced memory composition, namely \textit{Surprise}, \textit{Max. Loss} and \textit{MoF} (see Figure \ref{fig:composition}), benefit from higher memory capacities. Larger memory helps to avoid overwriting elements of past tasks, which counteracts imbalances in the composition of the memory. 





\begin{figure*}[htbp]
  \centering
  \vspace{-0.25cm}   
  \includegraphics[width=0.95\textwidth]{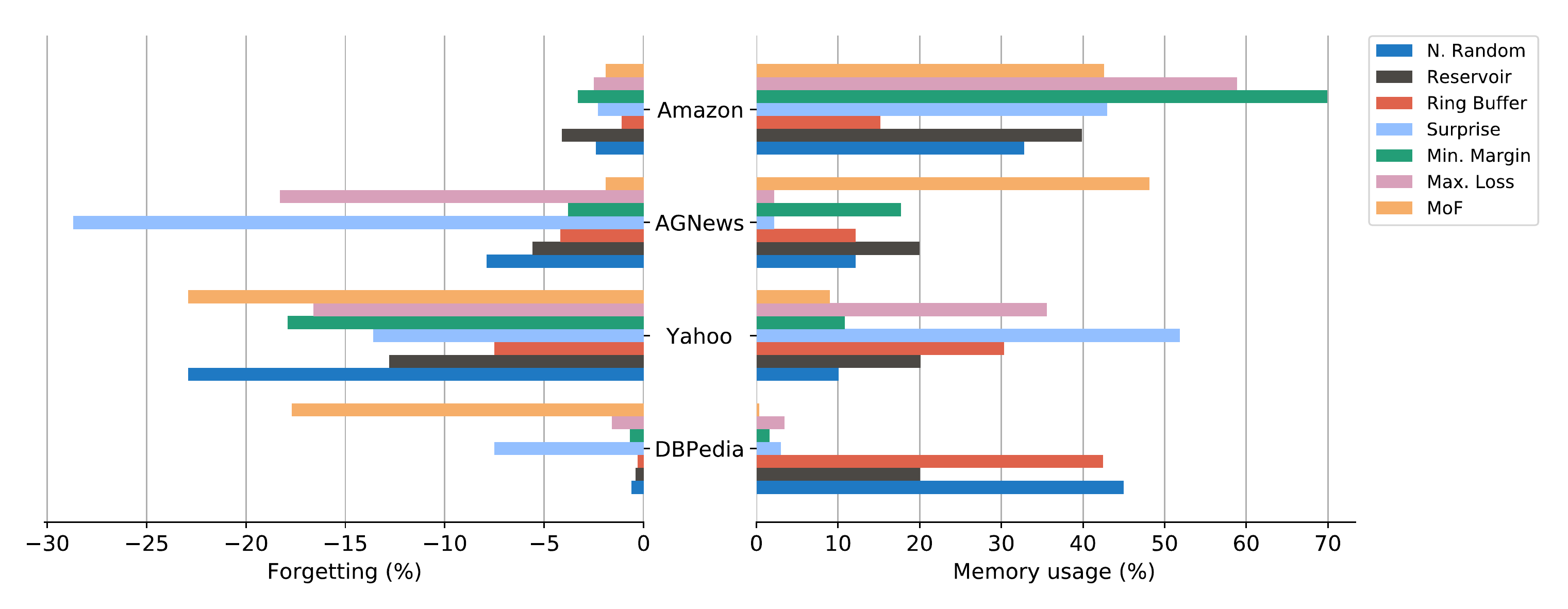}
  \vspace{-0.25cm}
  \caption{Double bar graph contrasting the percentages of forgetting and memory usage per task for all the population methods. Forgetting is computed by the difference between the current and previous model performance.}
  \label{fig:forgetting}
  \vspace{-0.35cm}   
\end{figure*}

\paragraph{Forgetting and Memory Usage.} 
To better understand why some methods perform worse, we compare the model forgetting and memory usage of text classification task - order (ii). Forgetting is the difference between a task's final performance and the initial performance. Memory usage is the percentage of items in memory (non-normalized) belonging to a task.

Figure~\ref{fig:forgetting} shows a direct relationship between a high forgetting percentage and few elements in memory. This is the main reason why the \textit{Surprise}, \textit{Max. Loss} and \textit{MoF} obtain the worst performance at 10\% memory.
However, there are some exceptions. \textit{Surprise} and \textit{Max. Loss} have many elements of the \textit{Yahoo} dataset, but forgetting is also high. We hypothesize those methods store examples that are not representative of the task's global distribution, resulting in a possible underfitting of the model.

Interestingly, Figure~\ref{fig:forgetting} shows that \textit{Reservoir} balances the number of samples in terms of tasks, which may be why this method surpass all others. Meanwhile, \textit{Ring Buffer} gets lower performance by balancing memory in terms of classes (Figure~\ref{fig:composition}), suggesting it is not the ideal way to fill the memory.

\paragraph{Influence of Resulting Memory on Local Adaptation} 

\begin{figure}
  \centering
  \includegraphics[width=0.45\textwidth]{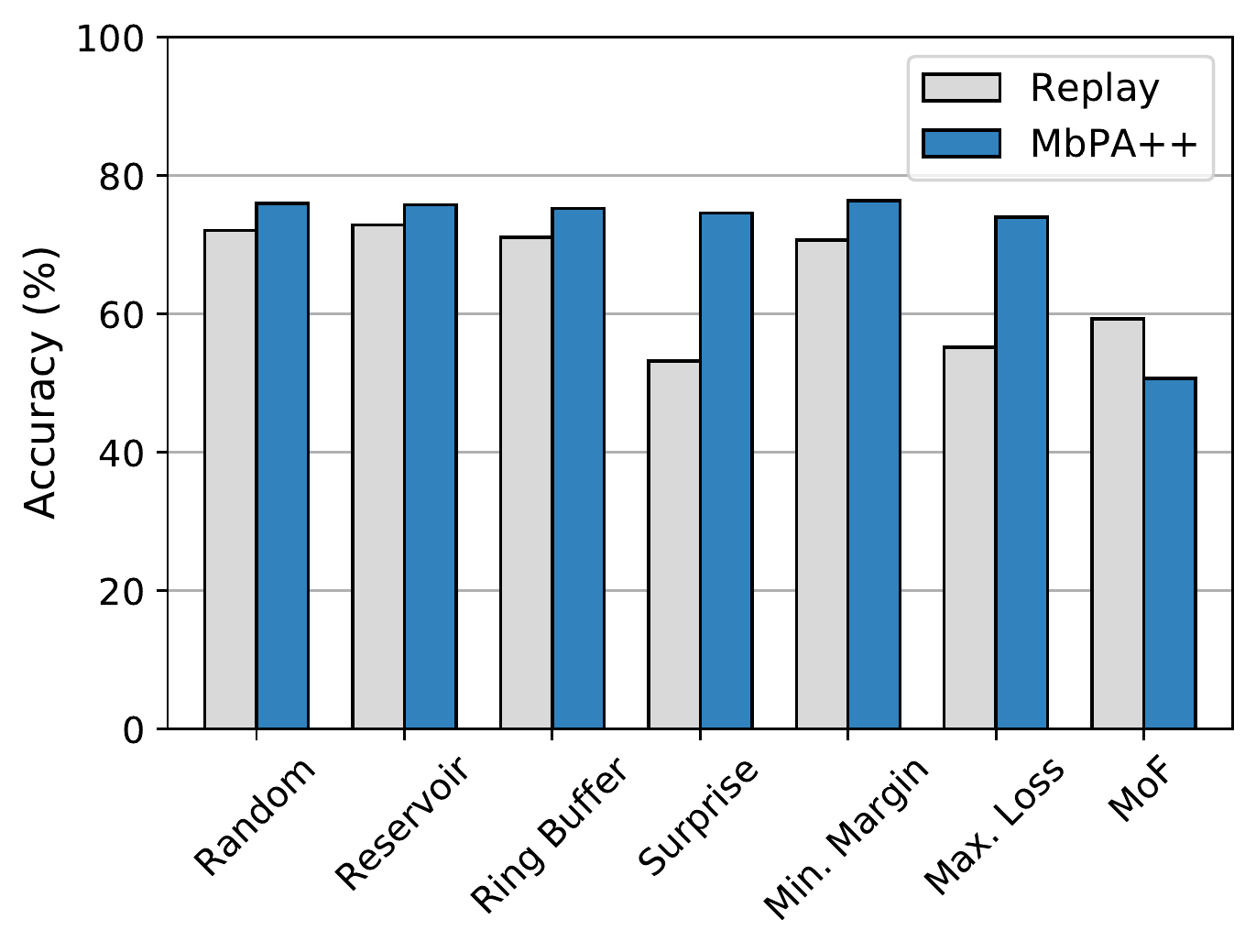}
  \vspace{-0.25cm}
  \caption{Influence of memory population methods when performing local adaptation to the replay model.}
  \label{fig:mbpa}
  \vspace{-0.35cm}   
\end{figure}

As mentioned in Section~\ref{sec:lll}, \citet{NEURIPS2019_f8d2e80c} proposed the MbPA++ model, which is a replay model with an additional local adaptation step during inference. We analyze how the resulting memory influences the local adaptation process of the text classification tasks - order (ii).

Figure~\ref{fig:mbpa} shows that the resulting memories of \textit{Surprise} and \textit{Max. Loss} methods benefit from local adaptation. We hypothesize that this is due to the criteria of these methods. Intuitively, the memory samples hard examples, which might be beneficial for local adaptation but not for replay, potentially leading to overall poor performance. Relative to the other methods, there is no significant increase in performance by applying local adaptation. This could be because the model has already reached the upper bound performance. Lastly, \textit{MoF} suffers from local adaptation, likely due to its suboptimal representations derived from \texttt{[CLS]} tokens.

\section{Conclusion}
In this work, we studied memory population methods for episodic memory in the context of lifelong language learning.
Our empirical analysis shows that simple methods such as Naive Random and Reservoir are the best choice for text classification and question answering because they randomly sample the global distribution. However, in the case of question answering, a balanced memory in terms of tasks leads to better results.

\section*{Acknowledgements}
This work was supported by the European Research Council Advanced Grant 788506, the National Center for Artificial Intelligence CENIA FB210017 - Basal ANID, and Vicerrectoría de Investigación de la Pontificia Universidad Católica de Chile - Concurso Puente 2021.

\bibliography{anthology,custom}
\bibliographystyle{acl_natbib}










\end{document}